\newcommand {\esiste} {\exists}

\newcommand {\tc} {\mid}

\newcommand{\alc}{\mathcal{ALC}}

\newcommand{\el}{\mathcal{EL}^{\bot}}
\newcommand{\elPP}{\mathcal{EL}^{++}}

\newcommand{\be}{\begin{enumerate}}
\newcommand{\ee}{\end{enumerate}}

\newcommand{\hide}[1]{}

 \hide{
{\qed \end{trivlist}}

\newenvironment{definition}
{\begin{defi} \rm}{\qed \end{defi}}

\newenvironment {proofof}[2]
{\begin{trivlist} \item[] {\bf Proof of #1~\protshoiqect{\ref{#2}}.}}%
{\qed \end{trivlist}}

\newenvironment{example}
{\begin{exa} \rm}{\qed \end{exa}}

}

\def \cases{\left \{\begin{array}{l}}
\def \endcases{\end{array}\right .}

\newcommand {\bes} {\begin{description}}
\newcommand{\ens} {\end{description}}
\newcommand {\la} {\langle}
\newcommand {\ra} {\rangle}

\newcommand {\beq} {\begin{quote}}
\newcommand {\enq} {\end{quote}}
\newcommand {\bit} {\begin{itemize}}
\newcommand {\enit} {\end{itemize}}

\newcommand {\bottom} {\bot}
\newcommand {\Unt} {{\cal U}}

\newenvironment{pozz}{\color{black}}{\color{black}}

\documentclass{llncs}
\usepackage{times}
\usepackage{undertilde}
\usepackage{graphicx}
\usepackage{latexsym}
\usepackage{graphicx}

\usepackage{color}
\usepackage{amssymb}

\definecolor{pozz}{rgb}{0,0,0}

\begin{document}
\bibliographystyle{splncs03}

\title{{
Reasoning about actions with $\cal EL$ ontologies \\ with temporal answer sets}
}

\author{\vspace{-0.15cm}
Laura Giordano \inst{1} \and Alberto Martelli \inst{2}  \and Daniele Theseider Dupr\'e \inst{1}}



\institute{DISIT - Universit\`a del  Piemonte
Orientale, Alessandria, Italy  \\ \email{laura.giordano@uniupo.it}, \ \  \email{dtd@di.unipmn.it}
\and Dip. di Informatica - Universit\`a di Torino, Italy \\ \email{mrt@di.unito.it}  
\vspace{-0.6cm}
}


\maketitle

\begin{abstract}
We propose an approach based on Answer Set Programming for reasoning about actions with domain descriptions including ontological knowledge, expressed in the lightweight description logic $\el$. 
We consider a temporal action theory, which allows for non-deterministic actions and causal rules to deal with ramifications, and whose extensions are defined by temporal answer sets.
We provide conditions under which action consistency can be guaranteed with respect to an ontology, 
 by a polynomial encoding of an action theory extended with an $\el$ knowledge base (in normal form) into a temporal action theory. 

\end{abstract}

\section{Introduction}
The integration of description logics and action formalisms has gained a lot of interest in the past years
\cite{BaaderAAAI05,BaaderLPAR10,ChangJAR2012,CalvaneseAAAI14}.
In this paper we explore the combination of a temporal action logic \cite{GiordanoTPLP13a} and an $\el$ knowledge bases, with the aim of allowing 
reasoning about action execution in the presence of the constraints given by  an $\el$ ontology.

As usual in many formalisms integrating description logics and action languages \cite{BaaderAAAI05,Baader2010,ChangJAR2012,CalvaneseAAAI14},
we regard inclusions in the $\mathit{KB}$ as state constraints of the action theory, which
we expect to be satisfied in the state resulting after action execution.
In the literature of reasoning about actions it is well known that causal laws and their interplay with domain constraints
are crucial for solving the ramification problem  \cite{McCain95,Lin95,Thielscher97,Denecker98,GiordanoJLC00,Giunchiglia&al:2004}.
%
%
In case knowledge about the domain is expressed in a description logic,
the issue has been considered, e.g.,\ in \cite{BaaderLPAR10} where causal laws are used to ensure
the consistency  with the TBox 
of the resulting state, after action execution.
For instance, given a TBox containing $\exists  \mathit{Teaches}. \mathit{Course} \sqsubseteq  \mathit{Teacher}$,
and an ABox (i.e., a set of assertions on individuals) containing the assertion $\mathit{Course}(math)$,
an action which adds the assertion $ \mathit{Teaches(john, math)}$, without also adding $ \mathit{Teacher(john)}$,
will not give rise to a consistent next state with respect to the knowledge base.
The addition of the causal law
{\bf caused} $ \mathit{Teacher(john)}$ {\bf if} $ \mathit{Teaches(john, math)} \wedge  \mathit{Course(math)}$
would enforce, for instance, the above TBox inclusion to be satisfied in the resulting state.


The approach proposed by Baader et al. \cite{BaaderLPAR10} uses causal relationships to deal with the ramification problem
in an action formalism based on description logics, and
it exploits a semantics of actions and causal laws in the style of Winslett's \cite{Winslett88} and McCain and Turner's \cite{McCain95} fixpoint semantics.
In this paper, we aim at extending this approach 
to reason about actions with an $\el$ ontology with {\em temporal answer sets}.
Reasoning about actions with {temporal answer sets} has been proposed in \cite{CILC2010_actions,GiordanoTPLP13a} by defining a temporal logic programming language for reasoning about {\em complex actions} and {\em infinite computations}. 
This action language, besides the usual LTL operators, allows for general Dynamic Linear
Time Temporal Logic (DLTL) formulas to be  included
in domain descriptions to constrain the space of possible extensions.
In \cite{GiordanoTPLP13a} a notion of Temporal Answer Set for domain descriptions is introduced,
as a generalization of the usual notion of Answer Set, and a translation of domain descriptions into
standard Answer Set Programming (ASP) is provided, by exploiting {\em bounded model checking techniques} 
for the verification of DLTL constraints, extending the approach developed by Helianko and Niemela \cite{Niemela03} for bounded LTL model checking with Stable Models.
An alternative ASP translation 
of this temporal action language has been investigated in \cite{GiordanoKR12,GiordanoJLC15},
by proposing an approach to bounded model checking which exploits the B\"{u}chi automaton construction while searching for a counterexample, with the aim of achieving completeness.
Our temporal action logic has been shown to be strongly related to extensions of the ${\cal A}$ language \cite{GelfondL98,Baral2000,Leone04,Babb2013,Giunchiglia&al:2004}. 
The temporal formalism is also related to the recent temporal extension of Clingo, {\em telingo} dealing with finite computations \cite{telingo2019}. 

The paper studies {\em extended temporal action theories}, combining the temporal action logic mentioned above with an $\el$ knowledge base.
It is shown that, for $\el$ knowledge bases in normal form, the consistency of the action theory extensions with the ontology can be verified by 
adding to the action theory a set of  causal laws and state constraints, by exploiting a fragment of the materialization calculus by Kr\"otzsch \cite{KrotzschJelia2010}. 
Furthermore, sufficient conditions on the action theory can be defined 
to repair the states resulting from action execution and guarantee consistency with TBox.
To this purpose, for each inclusion axiom in TBox, a suitable set of 
causal laws can be added to the action theory.
Our approach provides a polynomial encoding of an extended action theory, with an $\el$ knowledge base in normal form, into the language of the (DLTL) temporal action logic  studied  in \cite{GiordanoTPLP13a}.
The proof methods for this temporal action logic, based on bounded model checking, can then be exploited for reasoning about actions in the extended action theory.

A preliminary version of this work, which does not exploit a temporal action language, has been presented in CILC 2016 \cite{CILC2016}.

\section{The description logic $\el$}

We consider a fragment of the logic $\elPP$ \cite{rifel} that, for simplicity of presentation, does not include role inclusions and concrete domains.
The fragment, let us call it ${\el}$, includes the concept $\bot$ as well as nominals.

We let ${N_C}$ be a set of concept names, ${N_R}$ a set of role names
  and ${N_I}$ a set of individual names.
A concept in $\el$ is defined as follows:
\begin{quote}
 $C:= A \tc \top \tc \bot \tc  C \sqcap C \tc \exists r.C \tc \{a\}$
 \end{quote}
where $A \in N_C$ and $r \in {N_R}$.
Observe that complement, disjunction and universal restriction are not allowed in $\el$.

 A knowledge base $\mathit{K}$ is a pair $({\cal T}, {\cal A})$, where ${\cal T}$ is a TBox containing a finite set of  concept inclusions  $C_1 \sqsubseteq C_2$ and
${\cal A}$ is an ABox containing assertions of the form $C(a)$ and $r(a,b)$, with $C,C_1,C_2$ concepts, $r \in N_R$ and $a, b \in N_I$.


We will assume that the TBox is in normal form \cite{BaaderLTCS-Report-05-01}.
Let $BC_{\mathit{K}}$ be the smallest set of concepts containing $\top$, all the concept names occurring in $\mathit{K}$ and all nominals $\{a\}$, for any individual name $a$ occurring in $\mathit{K}$. An inclusion is in {\em normal form} if it has one of the following forms: $C_1 \sqsubseteq D$, \ \  $C_1 \sqcap C_2 \sqsubseteq D$,
\ \  $C_1 \sqsubseteq \exists r.C_2 $,  \ \ $\exists r.C_2 \sqsubseteq D$, where $C_1, C_2 \in BC_{\mathit{K}}$, and $D \in BC_{\mathit{K}} \cup \{ \bot\}$.
In \cite{BaaderLTCS-Report-05-01} it is shown that any TBox can be normalized in linear time, by introducing new concept and role names.

In the following we will denote with $N_{C,\mathit{K}}$, $N_{R,\mathit{K}}$ and $N_{I,\mathit{K}}$
the (finite) sets of concept names, role names and individual names occurring in $\mathit{K}$.
\begin{definition}[Interpretations and models]\label{semantics} An interpretation in $\el$
is any structure $( \Delta^I, \cdot^I )$ where:
 $ \Delta^I$ is a domain;
 $\cdot^I$ is an interpretation function that maps each
concept name $A$ to set $A^I \subseteq  \Delta^I$, each role name $r$
to  a binary relation $r^I \subseteq  \Delta^I \times  \Delta^I$,
and each individual name $a$ to an element $a^I \in  \Delta^I$.
Furthermore:  
    $\top^I=\Delta^I$, $\bot^I=\emptyset$;
     $\{a\}^I= \{a^I\}$;
    $(C \sqcap D)^I$= $C^I \cap D^I$;
    $(\esiste r.C)^I$= $\{x \in \Delta \tc \exists y \in C^I: (x,y) \in r^I \}$.
An interpretation $( \Delta^I, \cdot^I )$
satisfies an inclusion $C \sqsubseteq D$ if   $C^I \subseteq D^I$;
it satisfies an assertion $C(a)$ if $a^I \in C^I$;
it satisfies an assertion $r(a,b)$ if $(a^I,b^I) \in r^I$.

Given  a knowledge base $\mathit{K}=({\cal T}, {\cal A})$, an interpretation $( \Delta^I, \cdot^I )$ is a {\em model} of ${\cal T}$ if
$( \Delta^I, \cdot^I )$ satisfies   all  inclusions in ${\cal T}$;
$( \Delta^I, \cdot^I )$ is a {\em model} of $\mathit{K}$ if
$( \Delta^I, \cdot^I )$ satisfies   all  inclusions in ${\cal T}$ 
and all  assertions in ${\cal A}$.
${\cal A}$ is {\em consistent with ${\cal T}$} if there is a model of ${\cal T}$ satisfying all the assertions in ${\cal A}$.

\end{definition}

\section{Temporal Action Theories} \label{sec:temporalActionTheories}

In this paper we refer to the notion of the temporal action theory in \cite{GMS00}, a rule based fragment of which has been studied in
\cite{GiordanoTPLP13a,GiordanoJLC15}, 
which exploits the dynamic extension of LTL introduced by Henriksen and Thiagarajan, called Dynamic Linear Time Temporal
Logic (DLTL) \cite{Henriksen99}. In DLTL the next state modality is indexed by actions
and the until operator $ \Unt^\pi$
is indexed by a program $\pi$ which, as in PDL,
can be any regular expression built from atomic actions using
sequence ($;$), nondeterministic choice ($+$) and finite iteration ($*$).
The derived modalities $\la \pi \ra$ and $[\pi]$ can be defined as: $\la \pi \ra \alpha \equiv \top \Unt^\pi \alpha$ and
$[\pi] \alpha \equiv \neg \la \pi \ra \neg \alpha$. Similarly, $\bigcirc$ (next), $\Diamond$ and $\Box$ operators of LTL can be defined.
We let $\Sigma$ be a finite non-empty set of (atomic) actions and we refer to  \cite{GMS00,GiordanoTPLP13a} for the details concerning complex actions.

A \emph{domain description} $\Pi$ is a set of laws describing the effects of actions and
their executability preconditions. Atomic propositions describing the state of
the domain are called \emph{fluents}. Actions may have direct effects, described by
action laws, and indirect effects,
described by causal laws capturing the causal dependencies among fluents.

Let ${\cal L}$ be a first order language which includes a finite number of constants and variables, but no function symbol. Let ${\cal P}$ be the set of predicate symbols, $Var$ the set of variables
and $Cons$ the set of constant symbols.
We call {\em fluents}
atomic literals of the form $p(t_1,\ldots,t_n)$,
where, for each $i$, $t_i \in Var\cup Cons$.
A {\em simple fluent literal} (or {\em s-literal}) $l$ is
an atomic literals $p(t_1,\ldots,t_n)$
or its negation $\neg p(t_1,\ldots,t_n)$.
We denote by ${Lit_S}$ the set
of all simple fluent literals.
$Lit_T$ is the set of {\em temporal fluent literals}:
if $l \in Lit_S$, then $[a]l, \bigcirc l \in Lit_T$,
where $a$ is an action name (an atomic proposition, possibly containing variables), and
$[a]$ and $\bigcirc$ are the temporal operators introduced in the previous section.
Let $Lit = Lit_S \cup Lit_T \cup \{\bot,\top\}$,
where $\bot$ represents the inconsistency and $\top$ truth.
Given a (simple or temporal) fluent literal $l$, $not\;l$ represents the default negation of $l$.
A (simple or temporal) fluent literal possibly preceded by a default negation, will be called an
 {\em extended fluent literal}.

The laws are formulated as rules of a temporally extended logic programming language. Rules
have the form
\begin{equation} \label{box_law}
\Box( l_0  \leftarrow l_{1}, \ldots ,l_m , not\; l_{m+1} , \ldots ,  not\; l_n)
 \end{equation}
where the $l_i$'s are either simple fluent literals or temporal fluent literals,
with the following constraints:
(i)  If $l_0$ is a simple literal, then the body cannot contain temporal literals;
(ii) If $l_0=[a]l$, then the temporal literals in the body must have the form $[a]l'$;
(iii)  If $l_0= \bigcirc l$,   then the temporal literals in the body must have the form $\bigcirc l'$.
As usual in ASP, the rules with variables will be used as a shorthand for the set of their ground instances.

In the following we use of a notion of {\em state}: a set of ground fluent literals.
A state  is said to be  {\em consistent} if it is not the case that both $f$ and $\neg f$ belong to the state, or that $\bot$ belongs to the state.
A state is said to be {\em complete} if, for each fluent $f$, either $f$ or $\neg f$ belong to the state.
The execution of an action in a state may possibly change the values of fluents in the state through its direct and indirect effects,
thus giving rise to a new state.

While a law as (\ref{box_law}) can be applied in all states, a law 
 \begin{equation} \label{init_law}
l_0  \leftarrow l_{1}, \ldots ,l_m , not\; l_{m+1} , \ldots ,  not\; l_n
 \end{equation}
 which is not prefixed by $\Box$, only applies to the initial state.
 
A domain description can be defined as a pair $(\Pi, {\cal C})$, consisting of a set of laws $\Pi$ and a set of temporal constraints ${\cal C}$.
the following action laws describe the deterministic effect of the actions {\em shoot} and {\em load} for the Russian Turkey problem,
as well as the nondeterministic effect of action {\em spin}, after which the gun may be loaded or not:\\
$\Box( [shoot] \neg alive \leftarrow loaded)$\ \ \ \ \ \ \ \ \ \ \ \ \ \ \  \ \ \ \ \ \ \ \ \ \ \  
$\Box [load] loaded$\\
$\Box( [spin] loaded \leftarrow \; not \; [spin] \neg loaded) $\ \ \ \ \ \ 
$\Box( [spin] \neg loaded  \leftarrow \; not \; [spin] loaded) $
The following precondition laws:
 	$\Box ( [load]\bottom \leftarrow loaded) $
specifies that, if the gun is loaded, $load$ is not executable.
The program $ (\neg in\_sight?; wait)^* ;in\_sight?; load;shoot$
describes the behavior of the hunter who waits for a turkey until
it appears and, when it is in sight, loads the gun and shoots.
Actions  $in\_sight?$ and $\neg in\_sight?$ are test actions (we refer to \cite{GiordanoTPLP13a}).
If the constraint
 $\la (\neg in\_sight?; wait)^* ;in\_sight?; load;shoot \ra \top$
 is included in ${\cal C}$
then all the runs of the domain description which do not start with an execution of the given program will be
filtered out.  For instance, an extension in which in the initial state the turkey is not in sight  and
the hunter loads the gun and shoots is not allowed.

 As we will see later, this  temporal language is also well suited to describe causal dependencies among fluents
 as {\em static} and {\em dynamic causal laws}
similar to the ones in the action languages
${\cal K}$ \cite{Leone04} and
${\cal C}^+$ \cite{Giunchiglia&al:2004}. 


The semantics of a domain description has been defined based on 
{\em temporal answer sets}\cite{GiordanoTPLP13a}, which extend the notion of  {\em answer set} \cite{Gelfond}
to capture the  linear structure of temporal models. Let us shortly recall the main notions.
In the following, we consider the ground instantiations of the domain description $\Pi$, and we denote by
$\Sigma$ the set of all the ground instances of the action names in $\Pi$.



\subsection{Temporal answer sets}


A temporal interpretation is defined as a pair $(\sigma, S)$, where $\sigma \in \Sigma ^\omega$ is a sequence of  actions and $S$ is
a consistent set of ground literals of the form  $[a_1;\ldots;a_k]l$,
where $a_1\ldots  a_k$ is a prefix of $\sigma$ and $l$ is a ground simple fluent literal,
meaning that $l$ holds in the state obtained by executing $a_1\ldots  a_k$.
$S$ is {\em consistent} iff it is not the case that both $ [a_1;\ldots;a_k]l\in S$ and $ [a_1;\ldots;a_k]\neg l\in S$, for some $l$,
or $ [a_1;\ldots;a_k]\bot \in S$.
A temporal interpretation $(\sigma,S)$ is said to be {\em total} if either $ [a_1;\ldots;a_k]p\in S$ or $ [a_1;\ldots;a_k]\neg p \in S$,
for each $a_1\ldots a_k$ prefix of $\sigma$ and for each fluent name $p$.

We define the {\em satisfiability of a simple, temporal or extended literal $t$
in a partial temporal interpretation $(\sigma,S)$  in the state $a_1\ldots a_k$},
(written $(\sigma,S), a_1\ldots a_k \models t$) as:

$(\sigma,S), a_1\ldots a_k \models \top$, \hspace{5mm} $(\sigma,S), a_1\ldots a_k \not \models \bot$

$(\sigma,S), a_1\ldots a_k \models l$ \  \emph{iff} \  $ [a_1;\ldots;a_k]l \in S$, for $l$ s-literal

$(\sigma,S), a_1\ldots a_k \models [a] l$  \  \emph{iff}\   $ [a_1;\ldots;a_k;a]l \in S$ or

 \hspace{3.5cm} $a_1\ldots a_k, a$ is not a prefix of $\sigma$

$(\sigma,S), a_1\ldots a_k \models \bigcirc l$  \  \emph{iff}\  $ [a_1;\ldots;a_k;b]l \in S$,

 \hspace{3.5cm} where $a_1\ldots a_k b$ is a prefix of $\sigma$

$(\sigma,S), a_1\ldots a_k \models not\; l$ \  \emph{iff} \  $(\sigma,S), a_1\ldots a_k \not \models l$

 \hspace{1mm}

\noindent
The satisfiability of rule bodies 
in a temporal interpretation is defined as usual.
A rule $\Box( H \leftarrow Body)$ is satisfied in a temporal  interpretation $(\sigma,S)$ if,
for all action sequences $a_1\ldots a_k$ (including the empty action sequence $\varepsilon$),   $(\sigma,S), a_1\ldots a_k \models Body$
implies $(\sigma,S), a_1\ldots a_k \models H$.
%
A rule $H \leftarrow Body$ is satisfied in a partial temporal  interpretation $(\sigma,S)$ if,  $(\sigma,S), \varepsilon \models Body$
implies $(\sigma,S), \varepsilon \models H$. 


Let $\Pi$  be a set of rules over an action alphabet $\Sigma$, not containing default negation, and
let $\sigma \in \Sigma^\omega$.

\begin{definition}
A temporal interpretation $(\sigma,S)$ is a {\em temporal answer set of $\Pi$} if
$S$ is minimal (with respect to set inclusion) among the 
$S'$ such that $(\sigma,S')$
is a partial interpretation satisfying the rules in $\Pi$.


\end{definition}

To define answer sets of a program $\Pi$ containing negation,
given a temporal interpretation $(\sigma, S)$ over
$\sigma \in \Sigma^\omega$,  the {\em reduct, $\Pi^{(\sigma, S)}$, of  $\Pi$  relative to $(\sigma, S)$} is defined, by extending
Gelfond and Lifschitz' transform \cite{Gelfond&Lifschitz:98}, roughly speaking,
to compute a different reduct of $\Pi$ for each
prefix $a_1,\ldots,a_h$ of $\sigma$.

\begin{definition}
The {\em reduct, $\Pi^{(\sigma, S)}_{a_1,\ldots,a_h}$, of  $\Pi$ relative to $(\sigma, S)$ and to the prefix $a_1,\ldots,a_h$ of $\sigma$ },
is the set of all the rules
$[a_1;\ldots;a_h] ( H \leftarrow l_1, \ldots , l_m ) $
such that
$H \leftarrow l_1, \ldots , l_m $, $not\; l_{m+1} , \ldots , not\; l_n $
is in $\Pi$ and  $(\sigma, S), a_1,\ldots,a_h \not \models  l_i$, for all $i=m+1,\ldots,n$.

The {\em reduct $\Pi^{(\sigma, S)}$ of  $\Pi$  relative to $(\sigma, S)$} is the union of all reducts  $\Pi^{(\sigma, S)}_{a_1,\ldots,a_h}$
 for all prefixes $a_1,\ldots,a_h$ of $\sigma$.

\end{definition}
In definition above, we say that rule $[a_1;\ldots;a_h] ( H \leftarrow Body ) $ is satisfied in a temporal  interpretation $(\sigma,S)$ if 
$(\sigma,S), a_1\ldots a_k \models Body$
implies $(\sigma,S), a_1\ldots a_k \models H$.

\begin{definition}[\cite{GiordanoTPLP13a}]
A temporal interpretation $(\sigma, S)$ is an {\em answer set of $\Pi$} if
$(\sigma, S)$ is an answer set of the reduct $\Pi^{(\sigma, S)}$.
\end{definition}

 Observe that a partial interpretation $(\sigma,S)$ provides, for each prefix $a_1\ldots a_k$, a partial evaluation of fluents in the
state corresponding to that prefix.
The (partial) state $w^{(\sigma,S)}_{a_1\ldots a_k}$ obtained by the execution of the actions $a_1\ldots a_k$ in the sequence
can be defined as:
$w^{(\sigma,S)}_{a_1\ldots a_k} = \{ l :  [a_1;\ldots;a_k]l \in S\}$.

Although the answer sets of a domain description $\Pi$ are partial interpretations,
in some cases, e.g., when the initial state is complete and all fluents are inertial,
it is possible to guarantee that
the temporal answer sets of $\Pi$ are total.
%
The case of total temporal answer sets is of special interest 
as a total temporal answer set $(\sigma, S)$ can be regarded as a temporal model. 



\section{Combining Temporal Action Theories with $\el$ KBs} 

In this section we define a notion of {\em extended temporal action theory}, consisting of a temporal action theory plus an $\el$ knowledge base.
Our approach, following most of the proposals for reasoning about actions in DLs
\cite{BaaderAAAI05,Baader2010,ChangJAR2012,BaaderLPAR10,CalvaneseAAAI14} is to regard
the TBox as a set of state constraints, i.e. conditions that must be satisfied by any state of the world (in all possible extensions of the action theory), and the ABox as a set of constraints on all possible initial states.

\normalcolor


We want to regard DL assertions as fluents that may occur in our action laws as well as in the states of the action theory.
Given a (normalized) $\el$ knowledge base $\mathit{K}=({\cal T}, {\cal A})$, we require that: 
(a) for each (possibly complex) concept  $C$ occurring in $\mathit{K}$ 
there is a unary predicate $C \in {\cal P}$;
(b) for each role name $r\in N_{R,\mathit{K}}$ 
there is a binary predicate $r \in {\cal P}$; 
(c) the set of constants $Cons$ includes all the individual names occurring in the $\mathit{K}$, i.e. $N_{I,\mathit{K}} \subseteq Cons$.

Observe that if a complex concept such as $\exists r.C$ occurs in $\mathit{K}$,  there exists a predicate name $\exists r.C \in {\cal P}$ and,
for each $a \in N_{I,\mathit{KB}}$,  the fluent literals $(\exists r.C)(a)$ and $ \neg(\exists r.C)(a)$ belong to the set $Lit$ (we will still call such literals assertions).
Although classical negation is not allowed in $\el$, we use {\em explicit negation} \cite{Gelfond} to allow negative literals of the form $\neg C(a)$
in the action language (to allow for deleting an assertion from a state).

A simple literal in $Lit$ is said to be a {\em simple assertion} if it has the form $B(a)$ or $r(a,b)$ or $\neg B(a)$ or $\neg r(a,b)$, where $B\in BC_{\mathit{K}}$ is a base concept in $\mathit{K}$,
$r \in N_{R,\mathit{K}}$ and $a,b \in N_{I,\mathit{K}}$.
Observe that $\{a\}(c)$ and $\neg \{a\}(c)$ are simple assertions, while $(\exists r.C)(a)$ and $\neg(\exists r.C)(a)$ are non-simple assertions.
In order to deal with existential restrictions, 
in addition to the individual names $N_{I,\mathit{KB}}$ occurring in the $\mathit{KB}$ we introduce a finite set $Aux$ of auxiliary individual names,
as proposed in \cite{KrotzschJelia2010} to encode $\el$ inference in Datalog,
where $Aux$ contains a new individual name $aux^{A\sqsubseteq \exists r.B}$, for each inclusion $A\sqsubseteq \exists r.B$ occurring in the $\mathit{KB}$.
We further require that $Aux \subseteq Cons$.
\normalcolor


\begin{definition}[Extended action theory]
An {\em extended action theory} is a tuple $ (\mathit{K},\Pi,{\cal C})$, where:
 $\mathit{K}=({\cal T}, {\cal A})$ is an $\el$ knowledge base;
 $\Pi$ is a set of laws: action, causal, executability and initial state laws (see below);
 ${\cal C}$ is a set of temporal constraints.
\end{definition}


{\em Action laws} describe the immediate effects of actions. They have the form:
\hide{
$$\Box (  [a]l \leftarrow l_1, \ldots , l_m), $$
where $[a]l$ is a fluent literal and $l_1,\ldots,l_m$ are simple fluent literals. The meaning is that executing action
$a$ in a state in which conditions $ l_1, \ldots , l_m$
hold
causes the effect $l$ to hold.
To model nondeterministic actions, we allow action laws with disjunctive effects as follows:
}
%
\begin{equation} \label{a_law}
\Box ( [a] l_0  \leftarrow t_{1}, \ldots ,t_m , not\; t_{m+1} , \ldots ,  not\; t_n)
 \end{equation}
where $l_0$ is a simple fluent literal and the $t_i$'s are either simple fluent literals or temporal fluent literals of the form $[a]l$.
 Its meaning is that executing action
$a$ in a state in which the conditions $t_{1}, \ldots ,t_m$ hold and conditions $ t_{m+1}, \ldots , t_n$ do not hold
causes the effect $l_0$ to hold.
%
As an example, 
$\mathit{[Assign(c,x)]Teaches(x,c)} \leftarrow Course(c)$ means that executing the action of assigning a course $c$ to $x$ has the effect that $x$ teaches $c$. 
Non-deterministic effects of actions can be defined using default negation in the body of action laws, as for the  action {\em spin} in Section \ref{sec:temporalActionTheories}.

{\em Causal laws} describe indirect effects of actions. They have the form:
In $\Pi$ we allow two kinds of causal laws.
{\em  Static causal laws} have the form:
\begin{equation} \label{sc_law}
\Box ( l_0 \leftarrow l_{1}, \ldots ,l_m , not\; l_{m+1} , \ldots ,  not\; l_n)
\end{equation}
where the $l_i$'s  are simple fluent literals.
Their meaning is: if $ l_{1}, \ldots , l_m$ hold in a state
and $ l_{m+1}, \ldots , l_n$ do not hold in that state,
than  $l_0$ is caused to hold in that state.
{\em Dynamic causal laws} have the form:
\begin{equation} \label{dc_law}
\Box ( \bigcirc l_0  \leftarrow t_{1}, \ldots ,t_m , not\; t_{m
+1} , \ldots , not\; t_n)
\end{equation}
where $l_0$ is a simple fluent literal and the $t_i$'s are either simple fluent literals or temporal fluent literals
of the form $\bigcirc l_i$.
For instance,
$\mathit{\bigcirc Teacher(x) \leftarrow \bigcirc Teaches(x, y)}$ $\mathit{ \wedge Course(y)}$.
Observe that, differently from \cite{BaaderAAAI05,BaaderLPAR10}, we do not restrict direct and indirect effects of actions to be simple assertions.
{\em State constraints} that apply to the initial state or to all states can be obtained
when  $\bot$ occurs
in the head of initial state laws (\ref{is_law})
or static causal laws (\ref{sc_law}).


{\em Precondition laws} describe the executability conditions of actions. They have the form:
$\Box ( [a]\bottom \leftarrow l_1, \ldots ,l_m , not\; l_{m+1} , \ldots , not\; l_n) $
where $a\in \Sigma$ and  the $l_i$'s are simple fluent literals.
The meaning is that the execution of an action $a$ is not
possible in a state in which  $l_1,\ldots,l_{m}$ hold  and $ l_{m+1}, \ldots , l_n$ do not hold 

%

{\em Initial state laws}  are needed to  introduce conditions that have to hold in the initial state.
They have the form:
\begin{equation} \label{is_law}
l_0  \leftarrow l_{1}, \ldots , l_m , not\; l_{m+1} , \ldots ,  not\; l_n
\end{equation}
where the $l_i$'s are simple fluent literals.
Observe that initial state laws, unlike static causal laws, only apply to the initial state
as they are not prefixed by the $\Box$ modality.
As a special case, the initial state can be defined as a set of simple fluent literals. For instance,
the initial state
$\{ alive, \neg  in\_sight, \neg frightened \}$ is defined by
the initial state laws: $alive, \; \; \neg  in\_sight,  \; \; \neg frightened $.




%
Following Lifschitz \cite{Lifschitz:90} we call {\em frame fluents } the fluents to which the law of inertia applies.
Persistency of frame fluents from a state to the next one 
can be captured by introducing in $\Pi$ a set of causal laws, said {\em persistency laws} for all frame fluents $f$.
\begin{center}
$\Box (\bigcirc f \leftarrow  f , \; not \bigcirc\neg f)$\ \ \ \ \ \ \ \ \ \ \ \ \ 
$\Box (\bigcirc \neg f \leftarrow  \neg f , \; not \bigcirc f)$
\end{center}
meaning that, if $f$ holds in a state,
then $f$ will hold in the next state, unless its negation  $\neg f$ is caused to hold (and similarly for $\neg f$).
%
Persistency of a fluent is blocked by the execution of an action which causes the value of the fluent to change, or by a nondeterministic action which may cause it to change.
The persistency laws above play the role of {\em inertia rules} in ${\cal C}$ \cite{Giunchiglia&al:98},
${\cal C}+$ \cite{Giunchiglia&al:2004}
and  ${\cal K}$ \cite{Leone04}.


If $f$ is {\em non-frame} with respect to an action $a$,
$f$ is not expected to persist and may change its value when an action $\alpha$ is executed,  
either non-deterministically:
\begin{center}
$\Box (\bigcirc f \leftarrow not \bigcirc\neg f)$\ \ \ \ \ \ \ \ \ \ \ \ \ 
$\Box (\bigcirc \neg f \leftarrow not \bigcirc f)$
\end{center}
or by taking some default value (see \cite{GiordanoTPLP13a} for some examples).

%

In the following we assume that persistency laws and non-frame laws can be applied to simple assertions 
but not to non-simple ones  (such as $(\exists r.B)(x)$), whose value in a state (as we will see) is determined from the value of simple assertions. 
For simple assertions $A(c)$ in a domain description, one has to choose whether the  concept $A$ is frame or non-frame (so that either persistency laws or non-frame laws can be introduced).
In particular, we assume that all the nominals always correspond to frame fluents:  if $\{a\}(b)$ (respectively $\neg \{a\}(b)$) belongs to a state, it will persist to the next state unless it is cancelled  by the direct or indirect effects of an action.
ABox assertions may incompletely specify the initial state.
As we want to reason about 
states corresponding to $\el$ interpretations, we assume that the laws:
$ f \leftarrow not \neg f$ and
$ \neg f \leftarrow not  f$
for {\em completing the initial state}
are introduced in $\Pi$ for all  simple literals $f$ (including assertions with nominals). 
As shown in \cite{GiordanoTPLP13a}, the assumption of complete initial states, together with suitable conditions on the laws in $\Pi$,
gives rise to semantic interpretations (extensions) of the domain description in which all
states are complete. 
In particular, to guarantee that each reachable state in each extension of a domain description is complete for simple fluents, 
we assume that, either the fluent is frame, and persistency laws are introduced for it, or it is non-frame, and non-frame laws are introduced.
 Other literals, such as existential assertions, are not subject to this requirement
but, as we will see below, the value of existential assertions in a state will be determined from the value of simple assertions.
Under the condition above, starting from an initial state which is complete for simple fluents, all the reachable states are also complete for simple fluents. We call {\em well-defined} the domain descriptions satisfying this condition (and sometimes we will simply say that $\Pi$ is well-defined).

The third component ${\cal C}$ of a domain description $(K,\Pi, {\cal C})$
is the set  of {\em temporal constraints} in DLTL,
which allow general temporal conditions to be imposed on the executions of the domain description.
Their effect is that of restricting the space of the possible executions (or {\em extensions}).
For a domain description $D=(\Pi,\cal C)$, as introduced in \cite{GiordanoTPLP13a}, extensions are defined as follows.
\begin{definition}
An {\em extension of a well-defined domain domain description $D=(\Pi,\cal C)$ over $\Sigma$} is a (total) answer set $(\sigma, S)$  of $\Pi$  which satisfies
the constraints in ${\cal C}$. 
\end{definition}
In the next section we extend the notion of extension to a domain description $(K,\Pi,\cal C)$ with $K$ an ontology.

\section{Ontology axioms as state constraints}


Given an action theory $(\mathit{K},\Pi, {\cal C})$, where $\mathit{K}=({\cal T}, {\cal A})$, we define an extension of 
$(\mathit{K},\Pi, {\cal C})$ as an extension $(\sigma, S)$ of the action theory $(\Pi, {\cal C})$ satisfying all axioms of the ontology $K$.
Informally,
each state $w$ in the extension is required to correspond to an $\el$ interpretation and to satisfy all inclusion axioms in TBox ${\cal T}$. Additionally, the initial state must satisfy all assertions in the ABox ${\cal A}$.

%

To define the extensions of an action theory $(\mathit{K},\Pi, {\cal C})$,
we restrict to well-defined domain descriptions $(\mathit{K},\Pi, {\cal C})$, so that all states in an extension are complete for simple fluents (and for simple assertions). We prove that such states represent $\el$ interpretations on the language of $K$, provided an additional set of laws $\Pi_{{\cal L}(K)}$ is included in the action theory.  Next, we add to $\Pi$ another set $\Pi_{\cal T}$ of constraints, to guarantee that each state satisfies the inclusion axioms in ${\cal T}$. Finally, we add to $\Pi$ the set of laws $\Pi_{\cal A}$, to guarantee that all assertions in ${\cal A}$ are satisfied in the initial state.

Overall, this provides a transformation of the action theory $(\mathit{K},\Pi, {\cal C})$ into a new action theory $(\Pi \cup \Pi_K, {\cal C})$, by eliminating the ontology while introducing the set of static causal laws and consrtraints $\Pi_K =  \Pi_{{\cal L}(K)} \cup \Pi_{\cal T} \cup \Pi_{\cal A}$, intended to exclude those extensions which do not satisfy the axioms in $K$.


The set of domain constraints and causal laws in  $\Pi_{{\cal L}(K)}$ is intended to guarantee that any state $w$ of an extension 
respects the semantics of  DL concepts occurring in $\mathit{K}$.  The definition of $\Pi_{{\cal L}(K)}$ is based on a fragment of the materialization calculus for $\el$,
which provides a Datalog encoding of an $\el$ ontology. 
Here, the idea is that of regarding an assertion $C(a)$ in a state $w$ as the evidence that $a^I \in C^I$ in the corresponding interpretation.
%
%
Let $\Pi_{{\cal L}(K)}$ be the following set of causal laws:
\begin{quote}
(1) $\Box(\bot \leftarrow \bot(x))$ \ \ \ \ \ \ \ \ \ \ \ \ \ \ \ 
(2) $\Box( \top(x) \leftarrow )$  \ \ \ \ \ \ \ \ \ \ \ \ \ \ \ 
%
%
%
%
(3) $\Box( \{a\}(a) \leftarrow )$


(4)  $\Box (exists\_r\_B(x) \leftarrow r(x,y) \wedge B(y) )$ 

(5) $\Box ((\exists r.B)(x) \leftarrow exists\_r\_B(x))$ 

(6)  $\Box(\neg (\exists r.B)(x) \leftarrow not \; exists\_r\_B(x))$




%




(7) $\Box( \bot \leftarrow \{a\}(x) \wedge B(x) \wedge not \; B(a))$, \ \  for $x\neq a$ \\   
(8) $\Box(  \bot \leftarrow \{a\}(x) \wedge B(a) \wedge not \; B(x))$, \ \  for $x\neq a$ \\   
(9) $\Box(  \bot \leftarrow \{a\}(x) \wedge r(z,x) \wedge not \; r(z,a))$, \ \  for $x\neq a$



\end{quote}
for all $x,y \in \Delta$, $a \in N_{I,\mathit{K}}$,
$B\in BC_{\mathit{K}}$ (the base concepts occurring in $\mathit{K}$)
and $r\in N_{R,\mathit{K}}$ (the roles occurring in $\mathit{K}$).
Observe that the first constraint has the effect that a state $S$, in which the concept $\bot$ has an instance,
is made inconsistent.
Law (4) makes $exists\_r\_B(x)$ hold in any state in which there is a domain element $y$ such that $r(x,y)$ and $ B(y) $ hold
(where $exists\_r\_B$ is an additional auxiliary predicate for $B\in BC_{\mathit{K}}$ and $r\in N_{R,\mathit{K}}$).
Laws (5) and (6) guarantee that, for all $x\in \Delta$,  either $ (\exists r.B)(x)$ or $\neg (\exists r.B)(x)$ is contained in the state.
State constraints (7-9) are needed for the treatment of nominals and are related to materialization calculus rules (27-29) \cite{KrotzschJelia2010}.

Let $(\sigma, S)$ be an extension of the action theory $(\Pi \cup \Pi_{{\cal L}(K)}, {\cal C})$.
It can be proven that any state $w$ of  $(\sigma, S)$ represents an $\el$ interpretation.
Given a state $w$, let $w^{+}$ be the set of $\el$ assertions $C(a)$ ($r(a,b)$),
such that $C(a) \in w$ (resp., $r(a,b)\in w$).
Let  $w^{-}$ be the set of $\el$ assertions $C(a)$ ($r(a,b)$), such that $\neg C(a) \in w$ (resp., $\neg r(a,b)\in w$).  


\begin{proposition}
Let $(\sigma, S)$ be an extension of the action theory $(\Pi \cup \Pi_{{\cal L}(K)}, {\cal C})$
and let $w$ be a state of $(\sigma, S)$.
Then there is an interpretation $( \Delta^I, \cdot^I )$  that satisfies all the assertions in $w^{+}$
and falsifies all assertions in $w^{-}$
(and we say that 
$( \Delta^I, \cdot^I )$ {\em agrees with state $w$}).
\end{proposition}

As mentioned, we are interested in the states satisfying the TBox ${\cal T}$ of $K$.
%
%
%
Provided $\Pi$ is well-defined, for each extension $(\sigma, S)$ of the action theory $(\Pi \cup \Pi_{{\cal L}(K)}, {\cal C})$, 
any state $w$  is consistent and complete for all simple literals (and hence, by (5) and (6), for all assertions). 
We say that $w$ {\em satisfies the TBox} ${\cal T}$ if for all interpretations $( \Delta^I, \cdot^I )$ such that $( \Delta^I, \cdot^I )$ agrees with state $w$, $( \Delta^I, \cdot^I )$ is a model of ${\cal T}$. 

The requirement that each $w$ should satisfy the TBox ${\cal T}$ can be incorporated in the action theory through a set of constraints, that we call $\Pi_{\cal T}$, by exploiting the fact that the TBox ${\cal T}$ is in normal form.
$\Pi_{\cal T}$ contains the following state constraints:

\begin{quote}

 $\Box( \bot \leftarrow A(x) \wedge not \; D(x))$, \ for  each $A \sqsubseteq D$ in ${\cal T}$;

 $\Box( \bot \leftarrow A(x) \wedge B(x) \wedge not \;D(x))$, \  for each $A \sqcap B \sqsubseteq D$ in ${\cal T}$;

 $\Box( \bot \leftarrow A(x) \wedge not \;  (\exists r.B)(x))$, \ for each $A \sqsubseteq \exists r.B $ in ${\cal T}$;

 $\Box( \bot \leftarrow (\exists r.B)(x) \wedge not \; D(x) )$, \ for each $\exists r.B \sqsubseteq D$ in ${\cal T}$;

\end{quote}
where $A,B \in BC_{\mathit{K}}$, $D \in BC_{\mathit{K}} \cup \{\bot\}$ and $x \in  N_{I,\mathit{K}} \cup Aux$.
For $D= \bot$, the condition $not \;D(x)$ is omitted. 
The following proposition can be proved for a well-defined  $\Pi$.
\begin{proposition}
Let $(\sigma, S)$ be 
an extension of the action theory $(\Pi \cup \Pi_{{\cal L}(K)}, {\cal C})$
and let $w$ be a state of $(\sigma, S)$.
$w$ satisfies ${\cal T}$ iff $w$ satisfies all constraints in $\Pi_{\cal T}$. 
\end{proposition}
%
We can then add the constraints in  $\Pi_{\cal T}$ to an action theory $(\Pi \cup \Pi_{{\cal L}(K)}, {\cal C})$ to single out 
the extensions $(\sigma, S)$ 
whose states all satisfy the TBox ${\cal T}$.
In a similar way, we can restrict to answer sets $(\sigma, S)$ whose initial state $w^{(\sigma,S)}_{\varepsilon} $ satisfies all assertions in $\cal A$, by defining a set of initial state constraints as: 
$$\Pi_{\cal A}= \{ \bot \leftarrow not\; A(c)) \mid A(c) \in {\cal A} \} \cup \{ \bot \leftarrow not\; r(c,d)) \mid r(c,d) \in {\cal A} \} $$
We define the {\em extensions of the extended action theory $(\mathit{K},\Pi, {\cal C})$} as the extensions 
of the action theory $(\Pi \cup  \Pi_K, {\cal C})$, where $\Pi_K=  \Pi_{{\cal L}(K)} \cup \Pi_{\cal T} \cup \Pi_{\cal A}$.

Given this notion of extension of an action theory $(\mathit{K},\Pi, {\cal C})$, we can verify, in the temporal action logic, if a sequence of actions is executable from the initial state (executability problem) or whether any execution of an action sequence makes some property (e.g., some assertion) hold in all reachable states (temporal projection problem). 

%
Let us consider the example from the introduction.
\begin{example} \label{exa1}
Let $\mathit{K}=({\cal T},{\cal A})$ be a knowledge base such that
${\cal T}$=  $\{\exists \mathit{Teaches}. \mathit{Course} \sqsubseteq \mathit{Teacher}\}$ and
${\cal A}$=  \{ $\mathit{Person(john)}, \mathit{Course(cs1)}\}$.
We assume that all simple assertions, i.e., $\mathit{Person(x)}$, $\mathit{Teacher(x)}$,
$\mathit{Course(x)}, \mathit{Teaches(x,y)}$, are frame for all $x,y \in N_{I,\mathit{K}} \cup Aux$.
Let us consider a state $w_0$ where John does not teach any course and is not a teacher.
If an action $\mathit{Assign(cs1,john)}$ were executed in $w_0$, given a $\Pi$ containing the action law
$\mathit{[Assign(cs1,john)]}\mathit{Teaches(john,cs1)}$, 
the resulting state would contain {\em Teaches(john,cs1)} and $(\exists${\em Teaches. Course)(john)},
but would not contain not {\em Teacher(john)},
thus violating the state constraint 

$\; \mbox{\ \ \ \ \ }$ $\Box( \bot \leftarrow (\exists \mathit{Teaches}. \mathit{Course})(x) \wedge not \; \mathit{Teacher}(x) )$,

\noindent 
in $\Pi_{\cal T}$. 
In this case, 
 there is no extension of the action theory
in which action  $\mathit{Assign(cs1,}$ $\mathit{john)}$  can be executed in the initial state. 

As observed in \cite{BaaderLPAR10}, when this happens, the action specification can be regarded as being underspecified, as it is not able to capture the dependencies among fluents which occur in the TBox.
To guarantee that TBox is satisfied in the new state, causal laws are needed which allow the state to be {\em repaired}.
In the specific case, adding causal law
$\Box(  \mathit{Teacher(x)}  \leftarrow  \mathit{Teaches(x, y)} \wedge \mathit{Course(y)})$
to $\Pi$ would suffice to cause  $\mathit{Teacher(x)}$ in the resulting state, as an indirect effect of action $\mathit{Assign(cs1,john)}$.
\end{example}

\section{ Causal laws for repairing inconsistencies: 
sufficient conditions}
\label{Sec:Causal_Laws}

 Can we identify which and how many conditions are needed to guarantee that an action theory is able to repair the state, after executing an action, so to satisfy all inclusions of a (normalized) $\el$ TBox, when possible? Let us continue Example \ref{exa1}.

%


\begin{example} \label{exa2}
Consider the case when action {\em retire(john)} is executed in a state $w_1$ where  {\em Person(john), {Course}(cs1), Teaches(john,cs1)} and  {\em Teacher(john)} hold.
Suppose that the action law:
 $[\mathit{retire(john)}] \neg \mathit{Teacher(john)}$  is in $\Pi$.
Then, $ \mathit{\neg Teacher(john)}$ will belong to the new state (let us call it $w_2$), but $w_2$ will still contain the literals:
{\em{Course}(cs1), Teaches(john,cs1)}, which persist from the previous state (as {\em Course} and {\em Teaches} are frame fluents). Hence, $w_2$ would violate the TBox ${\cal T}$.
To avoid this, $\Pi$ should contain some causal law 
to repair the inconsistency, for instance,
$ \Box( \neg \mathit{Teaches(x,y)}  \leftarrow  \neg \mathit{Teacher(x)} \wedge \mathit{Course(y)})$. 
 By this causal law, when John retires he stops teaching all the courses he was teaching before.
 In particular, he stops teaching $cs1$. On the contrary, 
 $ \Box( \neg \mathit{Course(y)}  \leftarrow  \neg \mathit{Teacher(x)} \wedge  \mathit{Teaches(x,y)} )$ would be unintended. 

\end{example}

As we can see,  the causal laws needed to restore consistency when an action is executed can in essence be obtained from the
inclusions in the TBox and from their contrapositives, even though not all contrapositives are always wanted.  

In general, while defining a domain description, one has to choose which causal relationships are intended 
and which are not.
The choice depends on the domain and should not be done automatically,
but some sufficient conditions 
to restore the consistency of the resulting state (if any) with a TBox (in normal form) can be defined.

The case of a (normalized) inclusion $A \sqsubseteq B$, with $A,B\in N_C$, is relatively simple;
the execution of an action $\alpha$ with effect $A(c)$ (but not $B(c)$), in a state in which none of $A(c)$ and $B(c)$ holds,
would lead to a state which violates the constraints in the $\mathit{KB}$.
Similarly for an action $\beta$ with effect $\neg B(c)$.
Deleting $B(c)$ should cause $A(c)$ to be deleted as well, if we want the inclusion $A \sqsubseteq B$ to be satisfied.
Hence, to guarantee that the TBox is satisfied in the new state, for each inclusion  $A \sqsubseteq B$, two causal laws are needed:
$\Box(   B(x) \leftarrow  A(x))$ and $\Box(   \neg A(x) \leftarrow  \neg B(x))$.


For an axiom $A \sqcap B \sqsubseteq \bot$, consider the concrete case $pending \sqcap approved \sqsubseteq \bot$, representing mutually exclusive states of a claim in a process of dealing with an insurance claim, we expect the following causal laws to be included: 


$\Box(   \neg pending(x) \leftarrow  approved(x))$  \ \ \ \ \ \ \ \ \ 
$\Box(   \neg approved(x) \leftarrow  pending(x))$

%
%
%

\noindent
even though the second one is only useful if a claim can become pending again after having become
(temporarily) approved. For the general case, we have the following.

%
%
%
%
%
%
Let us define a set $\Pi_{ C({\cal T})}$ of causal laws associated with ${\cal T}$ as follows:
\begin{itemize}
\item
\noindent
For  $A \sqsubseteq B$ in ${\cal T}$: \  $\Box(    B(x) \leftarrow  A(x))$ \ and \ \ $\Box(    \neg A(x) \leftarrow  \neg B(x))$;

\item
\noindent
For $A \sqcap B \sqsubseteq D$: \  $\Box(    D(x) \leftarrow  A(x) \wedge  B(x))$ \ and at least one among\\
%
 $\Box(    \neg A(x) \leftarrow  \neg D(x) \wedge  B(x))$ \  and \ \ $\Box(    \neg B(x) \leftarrow  \neg D(x) \wedge  A(x))$;


\item
\noindent
For $A \sqsubseteq \exists r.B $: \ 
 $\Box(   r(x,aux^{A \sqsubseteq \exists r.B}) \leftarrow A(x))$, \  
 $\Box(  B(aux^{A \sqsubseteq \exists r.B}) \leftarrow A(x))$  \ and \\
 $\Box(   {\neg A(x)} \leftarrow \neg (\exists r.B )(x))$;


\item
\noindent
 For $\exists r.B \sqsubseteq A$ in ${\cal T}$: \    $\Box(    A(x) \leftarrow  (\exists r.B)(x)$, \ 
$\Box(    \neg (\exists r.B)(x) \leftarrow  \neg A(x))$ and 
at least one of: 
$\Box(    \neg r(x,y) \leftarrow  \neg A(x) \wedge  B(x))$ and
$\Box(    \neg B(x)  \leftarrow  \neg A(x) \wedge   r(x,y))$.



\end{itemize}

%

\begin{proposition}
Given a well-defined action theory  $( \Pi, {\cal C})$ and a TBox ${\cal T}$, any state $w$ of 
an extension $(\sigma, S)$ of $( \Pi \cup \Pi_{{\cal L}(K)} \cup \Pi_{C({\cal T})} \cup \Pi_{\cal A}, {\cal C})$ satisfies  ${\cal T}$. 
\end{proposition}
Observe that, for the case for $A \sqsubseteq \exists r.B $, from $r(x,aux^{A \sqsubseteq \exists r.B})$ and $B(aux^{A \sqsubseteq \exists r.B})$
$(\exists r.B)(x)$ is caused
by laws (4-5) in $\Pi_{{\cal L}(K)}$. 
While the causal laws in $\Pi_{ C({\cal T})}$ are sufficient to guarantee the consistency of a resulting state with TBox ${\cal T}$, one cannot exclude that some action effects are inconsistent with TBox and cannot be repaired  (e.g., an action with direct effects $A(c)$ and $\neg B(c)$, conflicting with an axiom $A \sqsubseteq B$ in ${\cal T}$). In such a case, the action would not be executable.

Notice that the encoding above of $\el$ TBox into a set of temporal laws only requires a polynomial number of laws to be added to the action theory (in the size of  $K$). 
Based on this mapping, the proof methods for our temporal action logic, which are based on the ASP encodings of bounded model checking  \cite{GiordanoKR12,GiordanoTPLP13a,GiordanoJLC15}, can be exploited for reasoning about actions in an action theory extended with an $\el$ knowledge base.

\section{Conclusions and Related Work}

In this paper we have proposed an approach for reasoning about actions by combining a temporal action logic in \cite{GiordanoTPLP13a}, whose semantics is based on a notion of temporal answer sets, and  an $\el$ ontology.
It is shown that, for $\el$ knowledge bases in normal form, the consistency of the action theory extensions with the ontology can be verified by 
adding to the action theory a set of  causal laws and state constraints, by exploiting a fragment of the materialization calculus by Kr\"otzsch \cite{KrotzschJelia2010}. 
Starting from the idea by Baader et al.   \cite{BaaderLPAR10}  that causal rules can be used to ensure the consistency of states with the TBox, we have defined sufficient conditions on the action theory 
to repair the states resulting from action execution and guarantee consistency with TBox.
For each inclusion axiom, a suitable set of 
causal laws has to be added.
Our approach provides a polynomial encoding of an extended action theory, including an $\el$ knowledge base in normal form, into the language of the (DLTL based) temporal action logic  studied  in \cite{GiordanoTPLP13a}.
The proof methods for this temporal action logic, based on ASP encodings of bounded model checking  \cite{GiordanoKR12,GiordanoTPLP13a,GiordanoJLC15}, can then be exploited for reasoning about actions in an extended action theory.
It would also be interesting,  for action domains with finite executions, to investigate whether the action theories in \cite{GiordanoTPLP13a} can be encoded in in {\em telingo} \cite{telingo2019}, and whether the optimized implementation of {\em telingo}  can be exploited for reasoning about action in our extended action theories.

%


Many of the proposals in the literature for combining DLs with action theories focus on expressive DLs. 
In their seminal work  \cite{BaaderAAAI05}, Baader et al. study the integration of action formalisms with expressive DLs, from $\alc$ to ${\cal ALCOIQ}$,
under Winslett's {\em possible models approach} (PMA) \cite{Winslett88},  
based on the assumption that TBox is acyclic and on the distinction between defined and primitive concepts
(i.e., concept names that are not defined in the TBox), where only primitive concepts are allowed in action effects.
They determine the complexity of the executability and projection problems 
and show that they get decidable fragments of the situation calculus.
Our semantics departs from PMA as causal laws are considered.
As \cite{Liu2006} and  \cite{BaaderLPAR10} we do not require the restriction to acyclic TBoxes and primitive concepts in postconditions.


The requirement of acyclic TBoxes is lifted in the work by Liu et al.\ \cite{Liu2006}, where an approach to the ramification problem is proposed which does not use causal relationships, but exploits {\em occlusion} to provide a specification of the predicates that can change through the execution of actions. The idea is to leave to the designer of an action domain the control of the ramification of the actions.

Similar considerations are at the basis of the approach by Baader et al.\ \cite{BaaderLPAR10} that, instead, exploits causal relationships
for modeling ramifications in an action language for ${\cal ALCO}$,
and defines its semantics in the style of McCain and Turner fixpoint semantics \cite{McCain95} (the action theory does not deal with non-deterministic effects of actions). It is shown that temporal projection is decidable and \textsc{ExpTime}-complete.
%
In this paper, following \cite{BaaderLPAR10}, we exploit causal laws for modeling ramifications in the context of a temporal action language for $\el$.
It allows for non-deterministic effects of actions and for the distinction between frame and non-frame fluents  \cite{Lifschitz:90} (which is strongly related to occlusion used in \cite{Liu2006}) based on 
the temporal logic.
We have also provided sufficient conditions for an action specification to be consistent with a normalized $\el$ $\mathit{KB}$.



Ahmetai et al. \cite{CalvaneseAAAI14} study the evolution of Graph Structured Data as a result of updates expressed in an action language. They provide decidability and complexity results for expressive DLs such as ${\cal ALCHOIQ}br$ (under finite satisfiability) as well as for variants of DL-{\em lite}. Complex actions including action sequence and conditional actions are considered.
Complex actions are considered as well in \cite{ChangJAR2012}, where an action formalism is introduced for a  family DLs, from ${\cal ALCO}$ to
${\cal ALCHOIQ}$, exploiting PDL program constructors to define complex actions. As in \cite{BaaderAAAI05}, the TBox is assumed to be acyclic.

In \cite{BagheriHaririJAIR13} Description Logic and Action Bases are introduced, where an initial Abox evolves over time
due to actions which have conditional effects. 
In \cite{CalvaneseIJCAI13} the approach is extended to allow for different notions of 
{\em repairing} of the resulting state, such as a maximal subset  consistent with the Tbox, or the
intersection of all such subsets. In this paper, we rely on causal laws for repairing states; selecting the
appropriate causal laws means acquiring more knowledge, and allows for a finer control on the resulting state.

Our semantics for actions, as many of the proposals in the literature, requires that a state provides a complete description of the world and is intended to represent an interpretation of the $\el$ knowledge base.
An alternative approach has been adopted in \cite{BagheriHaririJAIR13}, where a state can provide an incomplete specification of the world.
In our approach, an incomplete state could be represented as an epistemic state, which distinguish between what is known to be true (or to be false) and what is unknown. An epistemic extension of our action logic, based on temporal answer sets, has been developed in \cite{GiordanoJLC15}, and it can potentially be exploited for reasoning about actions with incomplete states also in presence of ontological knowledge. We leave the study of this case for future work. 

A mixed temporal and conditional approach to deal with causality in action theories, 
provides a natural formalization of concurrent actions and of the dependency (and independency) relations between actions,
has been studied in \cite{AIJ2004}. Whether this approach can be 
extended to domain descriptions also including knowledge from a DL ontology is also an interesting issue, which we leave for future work.



\medskip




\begin{thebibliography}{10}

\bibitem{CalvaneseAAAI14}
Ahmetaj, S., Calvanese, D., Ortiz, M., Simkus, M.: Managing change in
  graph-structured data using description logics. In: Proceedings of the
  Twenty-Eighth {AAAI} Conference on Artificial Intelligence. pp. 966--973
  (2014)

\bibitem{rifel}
Baader, F., Brandt, S., Lutz, C.: {Pushing the} $\mathcal{EL}$ {envelope}. In:
  Kaelbling, L., Saffiotti, A. (eds.) Proc. IJCAI 2005. pp. 364--369.
  Edinburgh, Scotland, UK (August 2005)

\bibitem{BaaderLTCS-Report-05-01}
Baader, F., Brandt, S., Lutz, C.: {Pushing the} $\mathcal{EL}$ {envelope}. In:
  LTCS-Report LTCS-05-01. Inst. for Theoretical Computer Science, TU Dresden
  (2005)

\bibitem{BaaderLPAR10}
Baader, F., Lippmann, M., Liu, H.: Using causal relationships to deal with the
  ramification problem in action formalisms based on description logics. In:
  LPAR-17. pp. 82--96 (2010)

\bibitem{BaaderAAAI05}
Baader, F., Lutz, C., Milicic, M., Sattler, U., Wolter, F.: Integrating
  description logics and action formalisms: First results. In: Proc. AAAI 2005.
  pp. 572--577 (2005)

\bibitem{Baader2010}
Baader, F., Liu, H., ul~Mehdi, A.: Verifying properties of infinite sequences
  of description logic actions. In: ECAI. pp. 53--58 (2010)

\bibitem{Babb2013}
Babb, J., Lee, J.: Cplus2asp: Computing action language $\mathcal{C}$+ in
  {Answer Set Programming}. In: Proc. Logic Programming and Nonmonotonic
  Reasoning, {LPNMR} 2013. pp. 122--134 (2013)

\bibitem{BagheriHaririJAIR13}
{Bagheri Hariri}, B., Calvanese, D., Montali, M., {De Giacomo}, G., {De
  Masellis}, R., Felli, P.: Description logic knowledge and action bases. J.
  Artif. Intell. Res.  46,  651--686 (2013)

\bibitem{Baral2000}
Baral, C., Gelfond, M.: Reasoning agents in dynamic domains. In: Logic-Based
  Artificial Intelligence, pp. 257--279 (2000)

\bibitem{telingo2019}
Cabalar, P., Kaminski, R., Morkisch, P., Schaub, T.: telingo = {ASP} + time.
  In: Logic Programming and Nonmonotonic Reasoning - 15th International
  Conference, {LPNMR} 2019, Philadelphia, PA, USA, June 3-7, 2019, Proceedings.
  Lecture Notes in Computer Science, vol. 11481, pp. 256--269. Springer (2019)

\bibitem{CalvaneseIJCAI13}
Calvanese, D., Kharlamov, E., Montali, M., Santoso, A., Zheleznyakov, D.:
  Verification of inconsistency-aware knowledge and action bases. In: Proc.
  {IJCAI} 2013 (2013)

\bibitem{ChangJAR2012}
Chang, L., Shi, Z., Gu, T., Zhao, L.: A family of dynamic description logics
  for representing and reasoning about actions. J. Autom. Reasoning  49(1),
  1--52 (2012)

\bibitem{Denecker98}
Denecker, M., {Theseider Dupr\'{e}}, D., {Van Belleghem}, K.: An inductive
  definitions approach to ramifications. Electronic Transactions on Artificial
  Intelligence  2,  25--97 (1998)

\bibitem{Leone04}
Eiter, T., Faber, W., Leone, N., Pfeifer, G., Polleres, A.: A logic programming
  approach to knowledge-state planning: Semantics and complexity. ACM
  Transactions on Computational Logic  5(2),  206--263 (2004)

\bibitem{Gelfond}
Gelfond, M.: Handbook of Knowledge Representation, chapter 7, Answer Sets.
  Elsevier (2007)

\bibitem{Gelfond&Lifschitz:98}
Gelfond, M., Lifschitz, V.: The stable model semantics for logic programming.
  In: Logic Programming, Proc. of the 5th Int. Conf. and Symposium. pp.
  1070--1080 (1988)

\bibitem{GelfondL98}
Gelfond, M., Lifschitz, V.: Action languages. Electron. Trans. Artif. Intell.
  2,  193--210 (1998)

\bibitem{GiordanoJLC00}
Giordano, L., Martelli, A., Schwind, C.: Ramification and causality in a modal
  action logic. J. Log. Comput.  10(5),  625--662 (2000)

\bibitem{GMS00}
Giordano, L., Martelli, A., Schwind, C.: Reasoning about actions in dynamic
  linear time temporal logic. The Logic Journal of the IGPL  9(2),  289--303
  (2001)

\bibitem{CILC2016}
Giordano, L., Martelli, A., Spiotta, M., {Theseider Dupr{\'{e}}}, D.: {ASP} for
  reasoning about actions with an \emph{EL\({}^{\bot}\)} knowledge base. In:
  Fiorentini, C., Momigliano, A. (eds.) Proceedings of the 31st Italian
  Conference on Computational Logic, Milano, Italy, June 20-22, 2016. {CEUR}
  Workshop Proceedings, vol. 1645, pp. 214--229. CEUR-WS.org (2016),
  \url{http://ceur-ws.org/Vol-1645/paper\_15.pdf}

\bibitem{CILC2010_actions}
Giordano, L., Martelli, A., {Theseider Dupr{\'{e}}}, D.: Reasoning about
  actions with temporal answer sets. In: Faber, W., Leone, N. (eds.)
  Proceedings of the 25th Italian Conference on Computational Logic, Rende,
  Italy, July 7-9, 2010. {CEUR} Workshop Proceedings, vol. 598. CEUR-WS.org
  (2010)

\bibitem{GiordanoKR12}
Giordano, L., Martelli, A., {Theseider Dupr{\'{e}}}, D.: Achieving completeness
  in bounded model checking of action theories in {ASP}. In: Brewka, G., Eiter,
  T., McIlraith, S.A. (eds.) Principles of Knowledge Representation and
  Reasoning: Proceedings of the Thirteenth International Conference, {KR} 2012,
  Rome, Italy, June 10-14, 2012. {AAAI} Press (2012),
  \url{http://www.aaai.org/ocs/index.php/KR/KR12/paper/view/4532}

\bibitem{GiordanoTPLP13a}
Giordano, L., Martelli, A., {Theseider Dupr{\'e}}, D.: Reasoning about actions
  with temporal answer sets. Theory and Practice of Logic Programming  13,
  201--225 (2013)

\bibitem{GiordanoJLC15}
Giordano, L., Martelli, A., {Theseider Dupr{\'{e}}}, D.: Achieving completeness
  in the verification of action theories by bounded model checking in {ASP}. J.
  Log. Comput.  25(6),  1307--1330 (2015),
  \url{https://doi.org/10.1093/logcom/ext067}

\bibitem{AIJ2004}
Giordano, L., Schwind, C.: Conditional logic of actions and causation. Artif.
  Intell.  157(1-2),  239--279 (2004)

\bibitem{Giunchiglia&al:2004}
Giunchiglia, E., Lee, J., Lifschitz, V., McCain, N., , Turner, H.: Nonmonotonic
  causal theories. Artificial Intelligence  153(1-2),  49--104 (2004)

\bibitem{Giunchiglia&al:98}
Giunchiglia, E., Lifschitz, V.: An action language based on causal explanation:
  Preliminary report. In: Proc. AAAI/IAAI 1998. pp. 623--630 (1998)

\bibitem{Niemela03}
Heljanko, K., Niemel\"a, I.: Bounded {LTL} model checking with stable models.
  Theory and Practice of Logic Programming  3(4-5),  519--550 (2003)

\bibitem{Henriksen99}
Henriksen, J., Thiagarajan, P.: Dynamic linear time temporal logic. Annals of
  Pure and Applied logic  96(1-3),  187--207 (1999)

\bibitem{KrotzschJelia2010}
Kr{\"{o}}tzsch, M.: Efficient inferencing for {OWL} {EL}. In: Proc. {JELIA}
  2010. pp. 234--246 (2010)

\bibitem{Lifschitz:90}
Lifschitz, V.: Frames in the space of situations. Artificial Intelligence  46,
  365--376 (1990)

\bibitem{Lin95}
Lin, F.: Embracing causality in specifying the indirect effects of actions. In:
  {IJCAI} 95, Montr{\'{e}}al Qu{\'{e}}bec, Canada, August 20-25 1995, 2
  Volumes. pp. 1985--1993 (1995)

\bibitem{Liu2006}
Liu, H., Lutz, C., Milicic, M., Wolter, F.: Reasoning about actions using
  description logics with general tboxes. In: Proc. {JELIA} 2006, Liverpool,
  UK. pp. 266--279 (2006)

\bibitem{McCain95}
McCain, N., Turner, H.: A causal theory of ramifications and qualifications.
  In: Proc. {IJCAI} 95. pp. 1978--1984 (1995)

\bibitem{Thielscher97}
Thielscher, M.: Ramification and causality. Artif. Intell.  89(1-2),  317--364
  (1997)

\bibitem{Winslett88}
Winslett, M.: Reasoning about action using a possible models approach. In:
  Proc. AAAI, St. Paul, MN, August 21-26, 1988. pp. 89--93 (1988)

\end{thebibliography}

\end{document}